\documentclass[lettersize,journal]{IEEEtran}
\usepackage{amsmath,amsfonts,amssymb}
\usepackage[linesnumbered,ruled,vlined]{algorithm2e}
\usepackage{array}
\usepackage[caption=false,font=normalsize,labelfont=sf,textfont=sf]{subfig}
\usepackage{textcomp}
\usepackage{xcolor}
\usepackage{stfloats}
\usepackage{url}
\usepackage{verbatim}
\usepackage{graphicx}
\usepackage{cite}
\usepackage{multirow}

\begin{document}
\title{Path Planning Based on a One-Shot Sampling Skeleton Map}

\author{Gabriel O. Flores-Aquino, Octavio Gutierrez-Frias, and Juan Irving Vasquez%
\thanks{Gabriel O. Flores-Aquino is with the Centro de Investigación en Matematicas (CIMAT), A.C. Guanajuato, Gto 36023, Mexico.}  
\thanks{Octavio Gutierrez-Frias is with the Unidad Profesional Interdisciplinaria en Ingeniería y Tecnologías Avanzadas (UPIITA), Instituto Politécnico Nacional (IPN), Mexico City 07340, Mexico.}%
\thanks{Juan Irving Vasquez is with the Centro de Innovación y Desarrollo Tecnológico en Cómputo (CIDETEC), Instituto Politécnico Nacional (IPN), Mexico City 07700, Mexico.}%
}


\markboth{Journal of \LaTeX\ Class Files,~Vol.~14, No.~8, August~2021}%
{Shell \MakeLowercase{\textit{et al.}}: A Sample Article Using IEEEtran.cls for IEEE Journals}


\maketitle

\begin{abstract}
Path planning algorithms fundamentally aim to compute collision-free paths, with many works focusing on finding the optimal distance path. However, for several applications, a more suitable approach is to balance response time, path safety, and path length. In this context, a skeleton map is a useful tool in graph-based schemes, as it provides an intrinsic representation of the free workspace. However, standard skeletonization algorithms are computationally expensive, as they are primarly oriented towards image processing tasks. We propose an efficient path-planning methodology that finds safe paths within an acceptable processing time. This methodology leverages a Deep Denoising Autoencoder (DDAE) based on the U-Net architecture to compute a skeletonized version of the navigation map, which we refer to as SkelUnet. The SkelUnet network facilitates exploration of the entire workspace through one-shot sampling (OSS), as opposed to the iterative or probabilistic sampling used by previous algorithms. SkelUnet is trained and tested on a dataset consisting of 12,500 two-dimensional dungeon maps. The motion planning methodology is evaluated in a simulation environment with an Unmanned Aerial Vehicle (UAV) in 250 previously unseen maps and assessed using several navigation metrics to quantify the navigability of the computed paths. The results demonstrate that using SkelUnet to construct the roadmap offers significant advantages, such as connecting all regions of free workspace, providing safer paths, and reducing processing time. These characteristics make this method particularly suitable for mobile robots in structured environments.
\end{abstract}


\section{Introduction}
\label{sec1}

\begin{figure}[htbp]
\begin{center}
\includegraphics[width=0.5\textwidth]{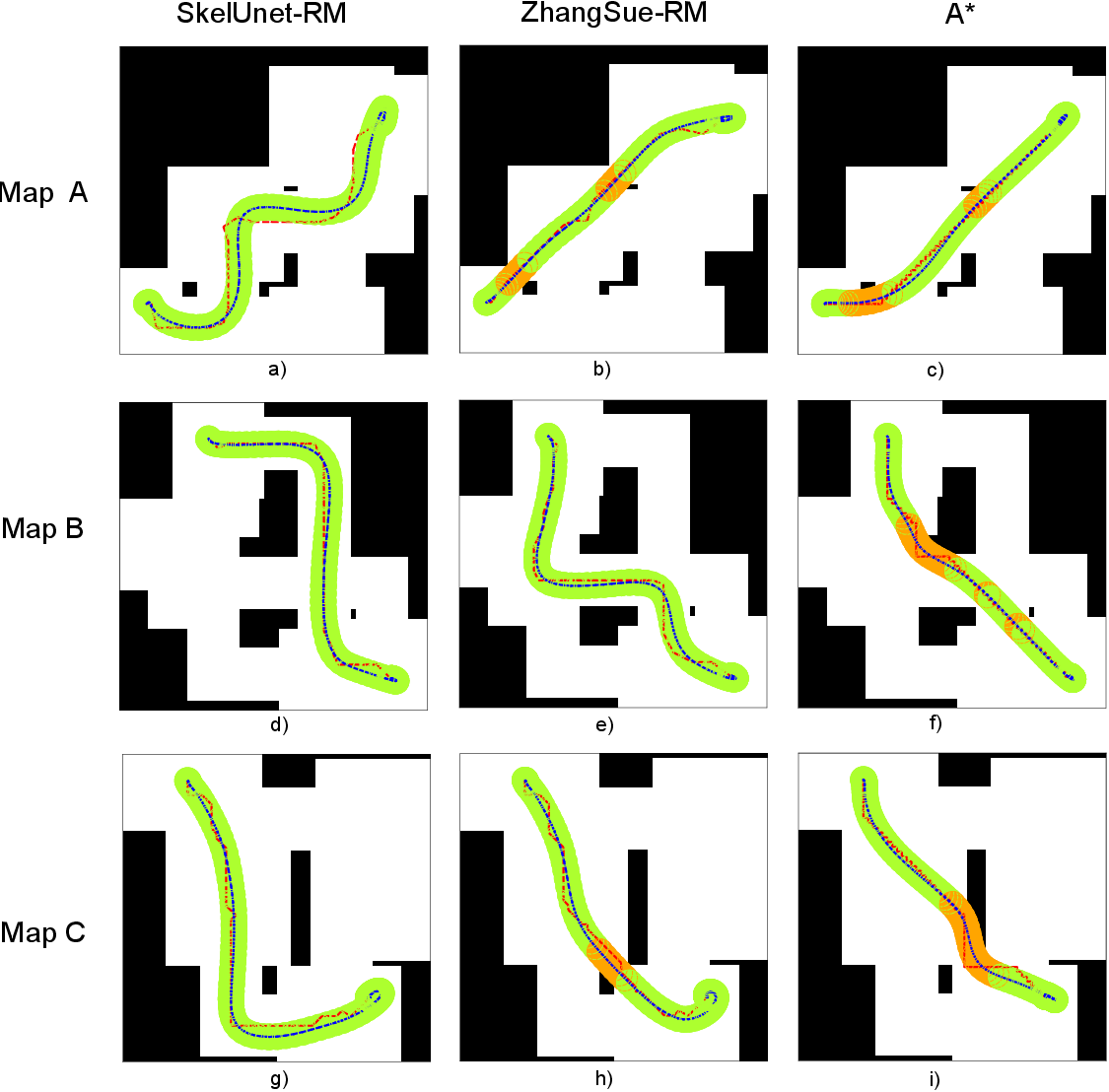}
\end{center}
\caption{Comparative path planning results across three indoor maps. Collision robot footprints are shown in green, while configuration within a collision radius are highlighted in orange. The planned path is indicated by a red line, and the executed trajectory is shown in blue. The first column is the trajectory using the skeleton obtained by SkelUnet the second row is the trajectory using the skeleton obtained by Zhang-Sue method and the third column is the trajectory computed with A* algorithm.}
\label{fig_comparativePath}
\end{figure}


Mobile robots designed for autonomous operation in structured environment face one of the field's persistent challenges: autonomous navigation \cite{corke2011robotics}\cite{HUANG2024124121}. A key component of autonomous navigation is motion planning, which aims to find a collision-free trajectory between an initial configuration, $x_{init}$, and a goal configuration, $x_{goal}$.  
In this context, a robot is defined as a rigid object $\mathcal{A}$, moving in a Euclidean space $\mathcal{W} \in \mathcal{R}^N$, called workspace. The configuration, $\mathcal{X}$, of any arbitrary object is the specification of the position of every point in this object relative to a workspace frame, $\mathcal{F}_{\mathcal{W}}$. A configuration $x$ of $\mathcal{A}$ is a specification of the position and orientation of the cartesian robot frame, $\mathcal{F}_{\mathcal{A}}$, relative to the global frame. 
The purpose of mapping the robot to the configuration space is to represent its geometry as single a point. In consequence, the path planning problem is reduced to find a sequence of robot configurations, $\mathcal{A}(x)$, that establish a collision-free path. A free path between two free configuration $(x_{init}, x_{goal})$ is a continuous map $\mathcal{S}:[0,1]\rightarrow \mathcal{X}_{free}$, where $\mathcal{X}_{free}=\{x\in \mathcal{X} / \mathcal{A}(x) \cap \left( \bigcup_{i=1}^{x} \mathcal{B}_{i}\right) \} $ and $\mathcal{B}$ is every obstacle in the workspace. Similarly, $\mathcal{X}$-obstacles can be described as $\mathcal{XB}=\{x\in\mathcal{X}/\mathcal{A}(x)\cap\mathcal{B}\neq \emptyset \}$. 

In many path planning problems, a map of the environment is available and the best strategy for finding a path is using the map information. The motion planning paradigm that uses maps is called map-based planning. A robot-friendly representation of a two-dimensional map is an occupancy grid map. In an occupancy grid map, we discretize the configuration space $\mathcal{X}$ into a rectangular grid $\mathcal{GX}$ as configuration in $\mathcal{GX}$ is labeled 0 if it lies in free space and 1 otherwise \cite{robotMotionPlanning}. If we have a map representation, the initial approach from a path planning perspective is to perform a cell search using a forward search algorithm, such as breadth-first, depth-first, Dijkstra’s algorithm, or A* \cite{robotMotionPlanning}. These algorithms are systematic methods that visit each reachable configuration in a finite amount of time and return a solution if one exists. This type of problem is known as discrete planning, and it faces the challenge of being computationally expensive because each planning exercise requires a new search, and the algorithms need to visit all reachable configurations. To address these issues, we can use a paradigm known as roadmap methods \cite{robotMotionPlanning}, where the map is utilized to build a representation of free configuration space. The roadmap is a type of topological graph that should provide an accurate and functional representation for navigation in free configuration space \cite{principlesOfRobotMotion}. It must contain selected information about the workspace and, additionally, serve as a helpful and reusable data structure for quickly finding collision-free paths.
There are many methods for building roadmaps, including visibility maps, deformation retracts, piecewise retracts, and silhouettes \cite{robotMotionPlanning}, to name a few. A roadmap is composed of vertices and edges. A vertex ($\mathcal{V}$) represents a reachable configuration while an edge ($\mathcal{E}$) is the connection between a pair of vertices. The roadmap has the aim of spreading out over the whole free configuration space $\mathcal{X}_{free}$ to be able to connect the initial vertex to a departure vertex and the goal vertex with an arrival vertex. Therefore, an adequate roadmap should capture the connectivity of the free configuration space in the form of a graph of one-dimensional curves.  Consequently, path planning is simplified to connecting the start and goal configurations to the roadmap and searching for a path in the graph $\mathcal{G}$ \cite{robotMotionPlanning}. 
    
\begin{figure}[tb]
\begin{center}
\includegraphics[scale=0.9]{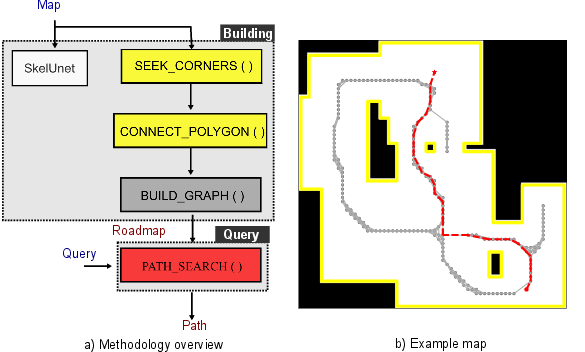}
\caption{SkelUnet-OSS planning scheme. The routine $\mathtt{SEEK\_CORNERS}$ found the corners inside the map, and the routine $\mathtt{CONNECT\_POLYGON}$ connects these corners in the correct order. In $\mathtt{BUILD\_GRAPH}$ we connect the vertices for the nearest neighborhoods and determine if an obstacle exist. The routine $\mathtt{PATH\_SEARCH}$ is responsible for finding a path within the graph.
}
\label{fig_methodologyPathPlanning}
\end{center}
\end{figure}

We propose a method that builds a roadmap, when the configuration space is two-dimensional with polygonal $\mathcal{X}$-obstacles, using a classic task in computer vision known as skeletonization \cite{Song_2021_ICCV}. From the perspective of digital image processing, skeletonization is a morphological operation \cite{szeliski2022computer}. The skeleton is defined as the locus of points where wavefronts propagating from a shape's contour meet and undergo cancellation. It serves as a central line of symmetry consisting of points equidistant from the pattern, providing a unique and invertible description, \cite{Blum}.Some characteristics that the skeleton must have are:
\begin{itemize}
\item [1)] End points must not be removed.
\item[2)] The connection must not be broken.
\item[3)] Excessive erosion must not be caused. 
\end{itemize}
Several methods exist for generating a skeleton; some examples include Voronoi diagrams, distance transformation \cite{robotMotionPlanning}, thinning \cite{saha2017skeletonization}, morphological operations \cite{maragos1986morphological}, and the Zhang-Suen method \cite{zhang1984fast}. Calculating the skeleton is an expensive and complicated task in terms of processing time, which is why deep learning methods have been proposed in recent years \cite{Song_2021_ICCV}, \cite{SkelNetOn_2019}. However, the effectiveness of these methods is closely tied to the configuration of the original image. Therefore, the algorithms are generally suitable only for specific data-sets, typically composed of objects such as animals or human shapes. Their application in path planning schemes remains an under-explored area.\par 

Our approach introduces a tiny neural network architecture (SkelUnet) capable of learning a free workspace representation from navigation maps and suitable for mobile robots. SkelUnet has the advantage of removes the lines that connect the rooms in crosswise, Figure \ref{fig_comparativeResultsSkeleton}. This characteristic improves the performance in path planning schemes. The path planning methodology SkelUnet-OSS is a forthright method suitable for being implemented in mobile robots, whose tasks are deployed in structured environments because its processing time is short and the graph is far from the obstacles, increasing the robot's safety, see Figure \ref{fig_comparativePath}. 
In addition, SkelUnet-OSS methodology is capable of performing the whole sampling process in one computing cycle and therefore reduces the computing cost. Additionally, to evaluate the paths obtained in terms of navigability we present the results of implementing the benchmarking metrics described in \cite{BenchmarkingMetricGroundNavigation}.\par

The paper is organized as follows. Section \ref{sec:related_work} describes the recent advances in the field. In section \ref{motionPlanningApproach}, we detail the path planning approach, and in section \ref{resultsSimulation}, we present the results. In section \ref{medialAxis}, we compare our method against the medial axis transform and in section \ref{conclusions} we talk about the conclusions and future work.

\section{Related work}
\label{sec:related_work}

Some works without a learning approach. In \cite{yang2007roadmap}, the authors propose an algorithm that builds a roadmap from a skeleton map and subsequently replaces the intersection edges with connected polygons. This change improves the transitions in crossing areas. The work also stands out the fact that in many cases the optimal distance path is not always the safest, so using some type of skeleton as a framework for the roadmap is a feasible option. Another recent work is presented in \cite{dong2018faster} where it is proposed to use the skeleton obtained by the \textsl{Wavefront} method \cite{robotMotionPlanning} as a growth guide for the RRT algorithm \cite{lavalle2001randomized}. From a machine learning perspective, some works implement neural networks to compute a roadmap, for example \cite{bourbakis1997path} where is used a single layer Kohonen neural network or \cite{Neural_RRT} where a deep convolutional neural network (CNN) along with A* algorithm is used to compute an optimal path. These works are focused in reduce processing time and they are not interested in the quality or navigability of the paths. Other works focus on ensuring that the skeleton represents navigable areas, such as \cite{ryu2019improved} and \cite{saha2017skeletonization}, where they modify the map before obtaining the skeleton, widen the obstacles considering the geometry of the robot. However this pre-processing may be unnecessary if we generate critical nodes to build the roadmap, a critical node is a key point that divide a complex path into sub-paths more manageable. This is the case of \cite{GeneratingCriticalNodesResnet50}, where CNN ResNet50 is used to compute a path. Works focused on UAVs are, for example, \cite{fixedWingUAV}, \cite{anEfficientMPAForUAVsInObstacleClutteredEnvironment}, and \cite{RapidAstarARobustPathPlanningSchemeForUAVs}.

Although the present work utilizes the previous idea of skeletonization for roadmap generation. It addresses specific limitations regarding computational efficiency and the topological quality of the resulting graphs. Unlike traditional methods that require geometric pre-processing to ensure safety \cite{ryu2019improved, saha2017skeletonization}, or deep learning approaches that rely on computationally heavy backbones like ResNet50 \cite{GeneratingCriticalNodesResnet50} often ignoring path quality in favor of speed, our proposal balances lightweight inference with high-quality roadmap construction.


\section{Path Planning with OSS Maps}
\label{motionPlanningApproach}

Our path planning methodology relies on the concept of graph-based planning, where a set of vertices in the workspace are connected to create a graph, subsequently, the graph is used to find a feasible path between the start and goal configurations. In Sampling-based Motion Planning (SBMP) algorithms, the vertices are the result of sampling either the workspace or the configuration space. In our case, the sampling is performed via a one-shot sampling using the SkelUnet network (see sub-section \ref{architectureModel}). The path planning methodology (SkelUnet-OSS) converts the generated vertices to a graph and resolves navigation queries. In the following sections, we present SkelUnet-OSS, comprising a construction stage and a query stage, detailed in subsections \ref{offline} and \ref{online} respectively. The general procedure is summarized in Figure \ref{fig_methodologyPathPlanning} and described in Algorithm \ref{algDDAEPP}. 

\subsection{SkelUnet}
\label{architectureModel}
\begin{figure*}
\centering
\includegraphics[width=0.85\textwidth]{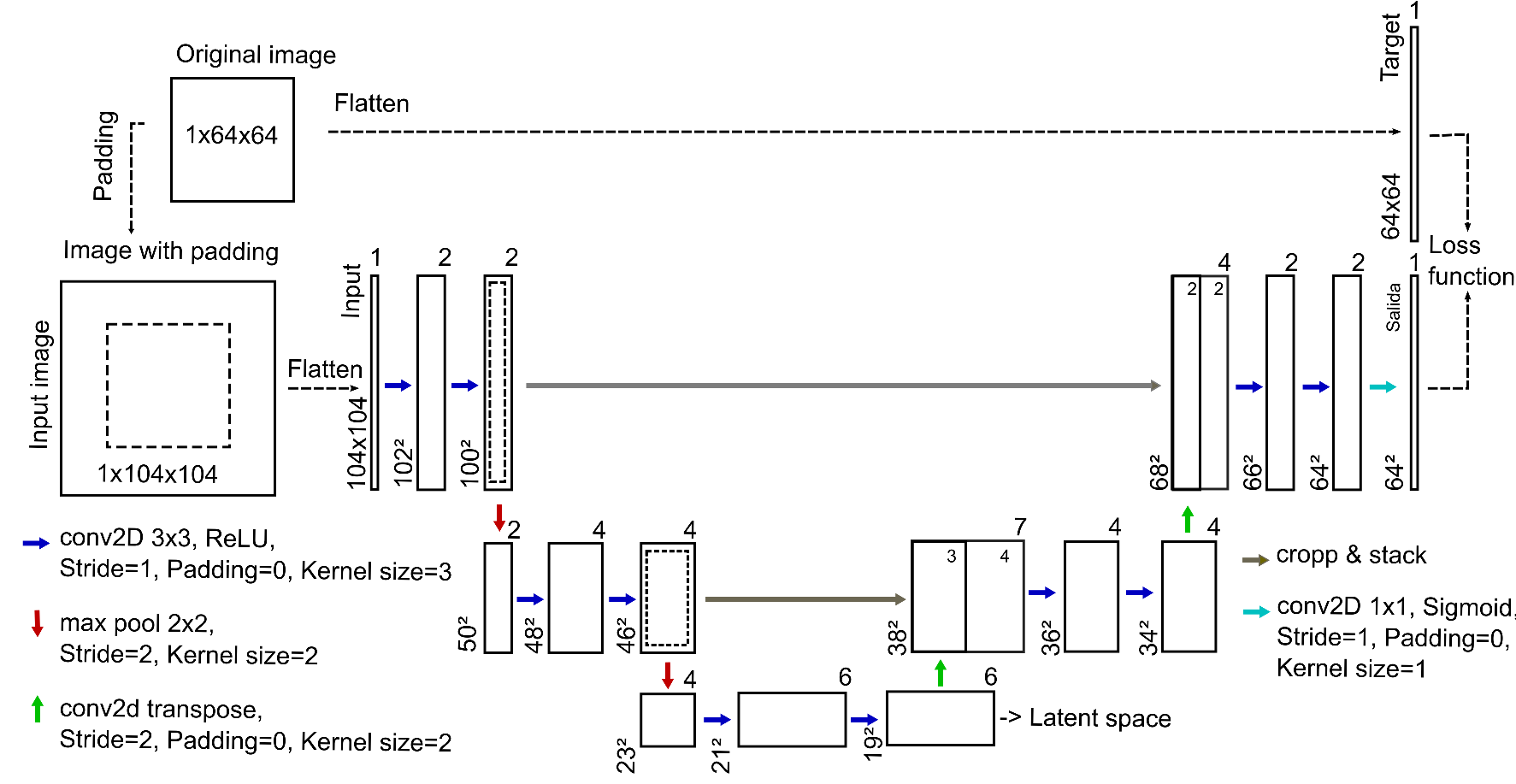}
\caption{SkelUnet architecture. The scheme shows the different operations performed by each layer and how it affects the size and depth of the image. The arithmetic of these operations can be consulted in \cite{dumoulin2016guide}.}
\label{figUNet}
\end{figure*}

U-Net architecture was originally designed for image segmentation and is formed by two neural networks in a U-shaped design: an encoder and a decoder. The encoder consist of convolutional and max pooling layers that progressively extract features i.e. encode the inputs to a latent representation; the decoder restores the original resolution using transposed convolutions \cite{UNET}. Based on the U-Net architecture, we propose SkelUnet as a way of sampling the workspace. Unlike the original U-Net, in our approach we consider the input image as a noisy version, and the target like a denoising output. SkelUnet, illustrated in Figure \ref{figUNet}, is designed to be trained in a supervised fashion, \cite{deepLearning}, where the input is a navigation map and the target is its skeletonized version. In this case, the target is obtained using the Zhang-Suen's method \cite{zhang1984fast}. Among the geometric methods, notable examples focus on either processing speed or solution accuracy, such as Voronoi-based skeletonization \cite{principlesOfRobotMotion}, thinning, and the Medial Axis Transform \cite{robotMotionPlanning}. Formally, SkelUnet is a convolutional deep denoising autoencoder (DDAE) composed of two parametric functions: 

\begin{equation}
h = f_{\theta}(x)=S_{f}(\text{Conv2d}(\theta, x))
\end{equation} for encoding and 
\begin{equation}
 g_{\phi}(h)=S_{g}( \text{UpConv2D}(\phi, h))
\end{equation}
 for decoding, where $\theta$ is the set of the convolution parameters. $\phi$ is the set of transposed convolution parameters, and $S_{f}$ with $S_{g}$ are activation functions\cite{bengio2013representation}. The rational for using an autoencoder-based architecture is motivated by the manifold hypothesis. This hypothesis posits that the probability distribution of the data of interest is highly concentrated in a reasonably small number of connected regions (manifolds), rather than being uniformly distributed over the entire space\cite{lecun2015deep}. Learning such a manifold is useful for sampling regions of interest, such as cluttered scenes or narrow passages, from a 2D workspace. In this paper, We use the original map representation, the workspace $\mathcal{W} \subset \mathbb{R}^2$, to feed the trained SkelUnet, obtaining a skeletonized version of the free workspace, $\mathcal{\widetilde{W}}_{free} \subset \mathbb{R}^2$. Note that for robots with higher degrees of freedom their free configuration space $\mathcal{X}_{free}$ may differ. Such a problem is out of the scope of this paper and it will be addressed in future work.

\subsection{Building Stage}
\label{offline}
We describe building stage in Algorithm \ref{algDDAEPP}. First, in lines 1 and 2 we build a geometric representation of the map. In line 1, the routine $\mathtt{SEEK\_CORNERS}$ finds all corners of the map and return their pixels coordinates in an array $\mathcal{C}$. In line 2, the routine $\mathtt{CONNECT\_POLYGON}$ connects each element of $\mathcal{C}$ and returns $n$ polygons in the set $\mathcal{XB}$. In line 3, we build a graph $\mathcal{G(\mathcal{V},\mathcal{E})}$ by calling the function $\mathtt{BUILD\_GRAPH}$. We need $\mathcal{XB}$ to verify the free transit between vertices. In line 6, we get a representation of the free workspace $\mathcal{\widetilde{W}}_{free}$. The propose of $\mathcal{\widetilde{W}}_{free}$ is get the set of vertices $\mathcal{V}\in \mathcal{W}_{free}$. In lines 7 to 14, we connect each vertex with its $k$ nearest neighbors. The notation $\overline{x_{near},x}$ refers to the edge $\mathcal{E}$ between this pair of vertices. Finally, we return the data structure $\mathcal{G}$.\\ 
\par
\subsection{Query Stage}
\label{online}
Query stage is described in the lines 18 to 23. In line 18, we make a query $\mathcal{Q}$ assigning $x_{I} \leftarrow q_{I}$ and $x_{G} \leftarrow q_{G}$. In lines 19 to 21, we connect the edge from initial configuration $x_{I}$ with departure vertex $x_{s}$ and the goal configuration $x_{G}$ with the arrival vertex $x_{f}$. Additionally, to add the current query to the roadmap, we need to verify that it has collision-free edges. Finally, we search for a feasible path $\mathcal{S}$ using the graph $\mathcal{G}$ as a roadmap, line 22. The algorithm returns the sequence of vertices and edges that resolves the current query. Note that the method works as a multi-query planner.  
\begin{algorithm}[tb]
\SetAlgoLined
\SetKwInOut{Input}{Input}
\SetKwInOut{Output}{Output}
\SetKwInOut{Parameters}{Parameters}
\Input{$\mathcal{W}$: Workspace; \\ $\Phi$: Trained 
SkelUnet;\\ $Q = \{q_{I}$, $q_{G}\}$: Query;}
\Output{$\mathcal{S}$: Path }
\Parameters{$k$: Number of neighborhoods;}
\SetKwProg{Fn}{$\mathtt{BUILD\_GRAPH}()$}{}{end}
\BlankLine

\tcc{\textbf{Building} stage:}

$\mathcal{C}\leftarrow\mathtt{SEEK\_CORNERS(\mathcal{W})};$

$\mathcal{XB}\leftarrow\mathtt{CONECT\_POLYGON}(\mathcal{W},\mathcal{C});$

$\mathcal{G}\leftarrow  \mathtt{BUILD\_GRAPH}(\Phi,\mathcal{W},\mathcal{XB},k)$;\\ 
\BlankLine
\Fn{}{
$\mathcal{G}\leftarrow\emptyset$;\\
$\mathcal{\widetilde{W}}_{free} \leftarrow \Phi(\mathcal{W})$;
$\mathcal{V} \leftarrow \widetilde{\mathcal{W}}_{free}\in\mathcal{W}_{free}$ \;
\For{$ x\in \mathcal{V}$}{
$\mathcal{X}_{nearest}\leftarrow\mathtt{NEAREST\_NEIGHBOR} (x,\mathcal{V},k)$;\\
	\For {$x_{near}\in \mathcal{X}_{nearest}$}
	{		
		\If{$\{\overline{x_{near},x} \cup \mathcal{XB}\}=\emptyset$}
		{
			$\mathcal{E}\leftarrow \overline{x_{near},x}$
		}
	}
}
$\mathcal{G}\leftarrow\{\mathcal{V,E}\}$;\\

\Return $\mathcal{G}$\;
}
\BlankLine
\tcc{\textbf{Query} stage:}

$x_{I},x_{G}\leftarrow Q\{q_{I},q_{G}\};$ 


\If{$(\{\overline{x_{I},x_{s}} \cup \mathcal{XB}\}=\emptyset)\;$ $\mathtt{and} \;$ $\{\overline{x_{G},x_{f}} \cup \mathcal{XB}\}=\emptyset)$}
{$\mathcal{G}\leftarrow \{x_{I},x_{G}\} \cup \mathcal{G}$}
$\mathcal{S}\leftarrow\mathtt{PATH\_SEARCH}(\mathcal{G},Q);$\\
\textbf{Return} $\mathcal{S}$
\caption{SkelUnet-OSS}
\label{algDDAEPP}
\end{algorithm}



\section{Results}
\label{resultsSimulation}

To validate the proposed path planning approach, the SkelUnet architecture was implemented, trained, and integrated into a motion planner. In the following sections, we describe the dataset used, the training of SkelUnet, the performance of the whole motion planning method in quadrotor navigation tasks, and provide a comparison against a previous skeletonization approach.

\subsection{Dataset}
\label{datasetForIndoorsMaps}

For training and testing SkelUnet, we utilize a dataset comprising 12,500 maps. We split the dataset into $80.0\%$ for training and $20.0\%$ for testing. Each map is a binary image of 64x64 pixels. The maps consist of hallways and rooms similar to classic dungeon maps, serving as a good representation of residential and office buildings. A complete review of the map generation process and dataset is available in \cite{GridMapGeneration}. For the training stage, we use the Zhang-Suen's method \cite{zhang1984fast} to generate the target images.

\subsection{SkelUnet training}
\label{skelUnetTraining}

In order to find a set of suitable parameters for the model, we conducted a grid-search across sixteen different configurations. We evaluated three different loss functions, three learning rated, and two optimization algorithms. The loss functions tested were Mean Squared Error (MSE), equation (\ref{eq:MSE}).

\begin{equation}
l_{n}=(x_{n}-y_{n})^2,    
\label{eq:MSE}
\end{equation}
Binary Cross Entropy (BCE), given by equation (\ref{eq:BCE}).
\begin{equation}
l_{n}=-w_{n}[y_{n}\cdot \log (x_{n})+(1-y_{n})\cdot \log (1-x_{n})],
\label{eq:BCE}
\end{equation}
and Weighted Sum (WS), given by equation (\ref{eq:WS}). 
\begin{equation}
l_{n}=\frac{(1-y_{n})\cdot(y_{n}-x_{n})^2}{c_{1}}+\frac{y_{n}\cdot(y_{n}-x_{n})^2}{1-c_{1}}
\label{eq:WS}
\end{equation}
with $c_{1}=0.8$. For all three functions, we compute the mean value to obtain the the final loss. The learning rates tested were $1e^{-3}$, $1e^{-4}$, and $1e^{-5}$; similarly the optimization algorithms tested were Adam and Stochastic Gradient Descent (SGD). For weight initialization, we used the He initialization. 

After passing the map through the autoencoder, we apply a threshold value between 0 and 1. This threshold is indicated with a subscript in Tables \ref{tableBestGridSearch} and \ref{tableF1Score}. SkelUnet was implemented in PyTorch. Training was conducted on a workstation equipped with an Intel i7-10700 CPU, an NVIDIA RTX 2070 super GPU, and 16GB of RAM. Training for 1000 epochs took about 2 hours and 50 minutes. To reduce the total training time, we selected the best three models after 1000 epochs; then we extended the training for 9000 epochs. The performance results and parameters for the best models are shown in Tables \ref{tableBestGridSearch} and \ref{tableF1Score}. The best performance is obtained with the parameters of the first row in the mentioned table.


\begin{center}
\begin{table}[tb]
\caption{Grid search results after 1000 epochs.}
\label{tableBestGridSearch}
\centering
\begin{tabular}{|l|l|l|l|l|l|l|}
\hline
\textbf{No.} & \textbf{\begin{tabular}[c]{@{}l@{}}LF\end{tabular}} & \textbf{\begin{tabular}[c]{@{}l@{}}LR\end{tabular}} & \textbf{Op} & \textbf{\begin{tabular}[c]{@{}l@{}}Tr\end{tabular}} & \textbf{Tst} & \textbf{$F1_{0.99}$} \\ \hline
1 & MSE & $1e^{-3}$ & Adam & 0.0118 & 0.0113 & \textbf{0.537} \\ \hline
2 & WS & $1e^{-3}$ & Adam & 0.0039 & 0.0075 & 0.521 \\ \hline
3 & BCE & $1e^{-3}$ & Adam & 0.0418 & 2.4407 & 0.487 \\ \hline
\end{tabular}
\end{table}
\end{center}

\begin{table*}[tb]
\caption{Results for metrics for validation dataset. For model 1 we obtained the best $F1$ score with $F1=0.8792$ and the best reconstruction loss $l=0.0020$ in model 3}
\label{tableF1Score}
\centering
\begin{tabular}{|l|l|l|l|l|l|l|l|}
\hline
\textbf{No.} & \textbf{TP} & \textbf{FP} & \textbf{TN} & \textbf{FN} & \textbf{Prc} & \textbf{Rcl} & \textbf{$F1_{0.5}$}   \\ \hline
1 & 94.264 & 9.176 & 3978.076 & 14.484 & 0.9014 & 0.8643 & \textbf{0.8792} \\ \hline
2 & 91.484 & 9.308 & 3977.944 & 17.264 & 0.9023 & 0.8390 & 0.8656 \\ \hline
3 & 99.132 & 21.056 & 3966.196 & 9.616 & 0.8203 & 0.9083 & 0.8602 \\ \hline
\end{tabular}
\end{table*}


\begin{figure}[tb]
\begin{center}
\includegraphics[width=0.45\textwidth]{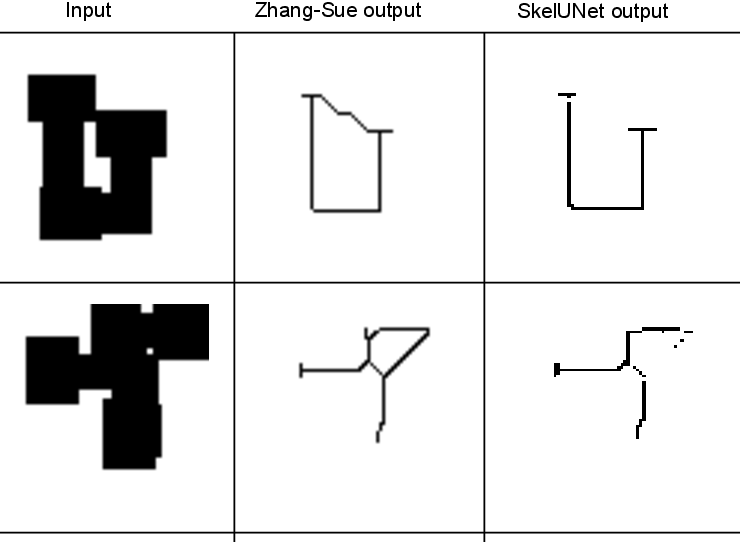}
\caption{Comparative skeletonization results using Zhang-Suen's method versus SkelUnet. The first column is the input, the second column shows the output from Zhang-Suen's method, and the third column shows the SkelUnet output.}
\label{fig_comparativeResultsSkeleton}
\end{center}
\end{figure}

\subsubsection{Results of SkelUnet architecture}

To evaluate the performance of the model, we used the F1-score (see Table \ref{tableF1Score}). This metric is particularly useful for binary data, where accuracy is often a biased indicator of performance due to class imbalance. For all the metrics, we consider skeleton pixels as the positive class. Thus, the precision ($Prc$) indicates, what proportion of the decoded skeleton is actually part of the target skeleton. The recall ($Rcl$) is the fraction of the skeleton that was detected. In other words, a low recall value means that SkelUnet tends to decode overly sparse skeletons (missing connections), whereas a low precision value implies that SkelUnet tends to decode a noisy or branched version of the skeleton (containing spurious artifacts). Values close to one indicate the highest performance.

For model 1, we obtained the best score with $F1_{0.5}=0.8792$, as shown in Table \ref{tableF1Score}. These are promising results if we take as a baseline the work of \cite{Song_2021_ICCV}, where the authors present a more extensive autoencoder model to skeletonize images containing different animal shapes and others forms; they report a best performance of $F1=0.6668$. In Figure \ref{fig_comparativeResultsSkeleton}, we show one of the advantages of using a learning-based method to compute the skeleton. We observe that SkelUnet removes unwanted connections that occur in diagonals. Since we want to use the skeleton map as a graph,  such kind of connections are not useful because they are generated between not connected rooms. Such a problem can be observed in the second row of the Figure \ref{fig_comparativeResultsSkeleton}.



\subsection{Robot Experiments} 

\begin{figure}
\centering
\includegraphics[width=\linewidth]{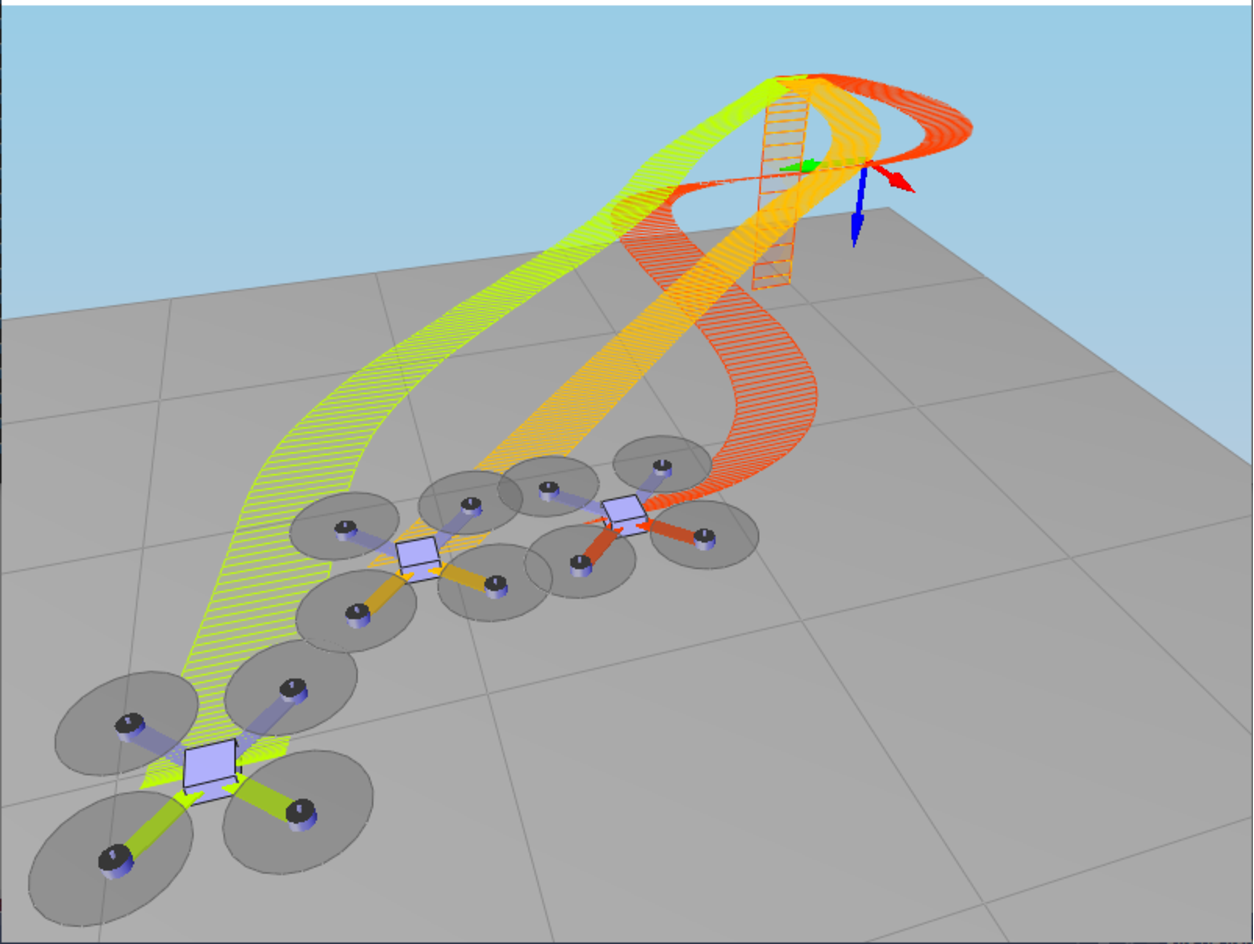}
\caption{Example of the environment for simulating drone dynamics. The quadrotor follows the paths obtained from SkelUnet in red, Zhang-Suen's method in green, and A* in yellow. This simulation validates dynamics; collision detection is performed using the dataset maps.}
\label{figFCND3drones}
\end{figure}

We used 250 maps and the navigation metrics described in \cite{BenchmarkingMetricGroundNavigation} to evaluate the paths obtained with SkelUnet-OSS. We applied the SkelUnet-OSS methodology using as input the skeleton derived from SkelUnet and Zhang-Suen. Additionally, we compared our results against a discrete planning method based on the A* algorithm. The benchmark metrics are shown in Table \ref{tablePathMetricsUNet}. Finally, SkelUnet-OSS results were tested in a simulation employing a quadrotor robot (see Figure \ref{fig_comparativePath}). Figure \ref{figFCND3drones} shows the environment to simulate drone dynamics. Order to check drone collision, first each path is simulated using drone dynamics, then the followed path and drone bounding box is projected to the 2D map and checked for collision. For the quadrotor, we use the z-axis exclusively for take-off and landing; path tracking is performed while maintaining a fixed altitude because our workspace is two-dimensional. The simulation environment implements a cascade PID controller and a second-order quadrotor dynamic model. Equation \eqref{eq:quadRotor1} describes the translational dynamics, while Equation \eqref{eq:quadRotor2} describes the rotational dynamics.

\begin{equation}
\begin{bmatrix} \ddot{x} \\ \ddot{y} \\ \ddot{z} \end{bmatrix} = R(t) \begin{bmatrix} 0 \\ 0 \\ c(t) \end{bmatrix} - \begin{bmatrix} 0 \\ 0 \\ g \end{bmatrix} \implies 
\begin{cases} 
\ddot{x} = c \cdot b_x \\ 
\ddot{y} = c \cdot b_y \\ 
\ddot{z} = c \cdot b_z - g 
\end{cases}
\label{eq:quadRotor1}
\end{equation}
Where $(b_x,b_y,b_z)$ correspond to the third column of the rotation matrix, and $c(t)$ the collective thrust.
\begin{equation}
\dot{R}(t) = R(t) \begin{bmatrix} 0 & -r(t) & q(t) \\ r(t) & 0 & -p(t) \\ -q(t) & p(t) & 0 \end{bmatrix}
\label{eq:quadRotor2}
\end{equation}
$p,q,r$ are the angular velocities of the body.
The physics and controller parameters are presented in table \ref{tabPhisycParmetersQuadRotor} and \ref{tabControlParmetersQuadRotor}. More details about the controller are available at \cite{schoellig2012feed}. 

\begin{table}[tb]
\caption{Quad-rotor physic parameters.}
 \label{tabPhisycParmetersQuadRotor}
 \centering
 \begin{tabular}{|ll|}
 \hline
 \textbf{Parameter}        & \textbf{Value} \\ \hline
 \multicolumn{1}{|l|}{Mass} & $0.5[kg]$            \\
 \multicolumn{1}{|l|}{Distance from vehicle origin to motors (\textbf{L})}    & $0.17[m]$           \\
 \multicolumn{1}{|l|}{Offset from center of mass on the x-axis (\textbf{cx})}   & $0.0[m]$              \\
 \multicolumn{1}{|l|}{Offset to center of mass on the y-axis (\textbf{cy})}   & $0.0[m]$              \\
 \multicolumn{1}{|l|}{Moments of inertia x-axis (Ixx)}  & $0.0023[kg\cdot m^2]$         \\
 \multicolumn{1}{|l|}{Moments of inertia y-axis (Iyy)}  & $0.0023[kg\cdot m^2]$         \\
 \multicolumn{1}{|l|}{Moments of inertia z-axis (Izz)}  & $0.0046[kg\cdot m^2]$        \\
 \hline
\end{tabular}
\end{table}

\begin{table}[]
\caption{Quad-rotor controller gains used in simulation.}
 \label{tabControlParmetersQuadRotor}
 \centering
\begin{tabular}{|ll|ll|ll|}
\hline
\multicolumn{2}{|c|}{\textbf{Position gains}} & \multicolumn{2}{c|}{\textbf{Velocity gains}}                & \multicolumn{2}{c|}{\textbf{Attitude gains}} \\ \hline
\multicolumn{1}{|l|}{kpPosXY}          & 1.0          & \multicolumn{1}{l|}{kpVelXY}                 & 1.5                  & \multicolumn{1}{l|}{kpBank}       & 13.5          \\ \hline
\multicolumn{1}{|l|}{kpPosZ}           & 2.5          & \multicolumn{1}{l|}{\multirow{2}{*}{kpVelZ}} & \multirow{2}{*}{7.0} & \multicolumn{1}{l|}{kpYaw}        & 2.5           \\ \cline{1-2} \cline{5-6} 
\multicolumn{1}{|l|}{kiPosZ}           & 2.0          & \multicolumn{1}{l|}{}                        &                      & \multicolumn{1}{l|}{kpPQR}        & 55,55,10      \\ \hline
\end{tabular}
\end{table}

\begin{table*}[tb]
\caption{Metric for paths obtained from skeletons using SkelUnet, Zhang-Sue's method and A*}
\label{tablePathMetricsUNet}
\centering
\begin{tabular}{|l|llll|llll|llll|}
\hline
& \multicolumn{4}{l|}{\textbf{SkelUnet}} & \multicolumn{4}{l|}{\textbf{Zhang-Sue}}                       & \multicolumn{4}{l}{\textbf{A*}} \\ \hline
& \textbf{Min.} & \textbf{Max.} & \textbf{Mean.} & \textbf{Std.} & \textbf{Min.} & \textbf{Max.} & \textbf{Mean.} & \textbf{Std.} & \textbf{Min.} & \textbf{Max.} & \textbf{Mean.} & \textbf{Std.} \\ \cline{1-12}
\textbf{DTCO.} & $2.71$ & $18.80$ & $8.32$ & $2.49$ & $2.81$ & $19.75$ & \textbf{8.78} & $2.80$ & $1.00$ & $22.65$ & $6.06$ & $3.30$ \\
\textbf{AV.} & $8.64$ & $30.81$ & $18.17$ & $3.91$ & $8.64$ & $31.45$ & \textbf{18.60} & $4.08$ & $8.64$ & $31.34$ & $17.54$ & $4.02$ \\
\textbf{Dsp.} & $0.00$ & $1.89$ & \textbf{0.25} & $0.30$ & $0.00$ & $2.22$ & $0.36$ & $0.47$ & $0.00$ & $4.00$ & $1.17$ & $0.78$ \\
\textbf{CD.} & $5.00$ & $48.00$ & $22.70$ & $7.10$ & $5.00$ & $53.00$ & \textbf{23.63} & $7.71$ & $5.00$ & $53.00$ & $22.01$ & $7.42$ \\
\textbf{Trts.} & $0.09$ & $1.00$ & $0.68$ & $0.17$ & $0.09$ & $1.00$ & $0.67$ & $0.16$ & $0.23$ & $1.00$ & \textbf{0.78} & $0.09$     \\     
\hline
\end{tabular}
\end{table*}

\begin{figure*}[tb]
\centering
\includegraphics[width=1.0\textwidth]{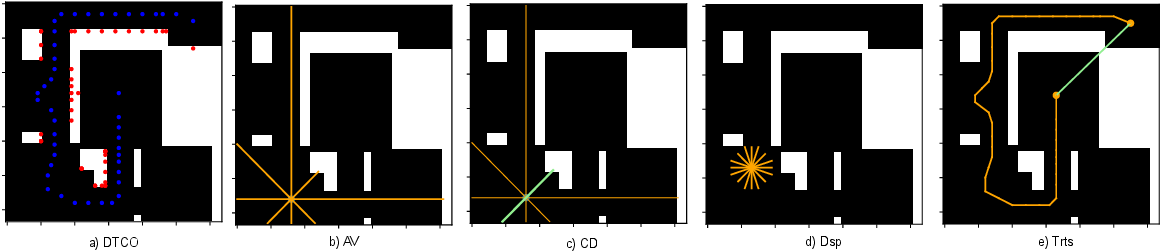}
\caption{Benchmark navigation ground metrics to evaluated the result paths. In this Figure the free workspace is highlighted in black.}
\label{figBenchmarkMetrics}
\end{figure*}

The trajectories obtained in simulation are evaluated using the metrics detailed in Figure \ref{figBenchmarkMetrics}. Distance to the Closest Obstacle (DTCO) is the distance to the nearest occupied region. For Average Visibility (AV), eight rays are traced, and the distance to the nearest obstacle is calculated in each direction. The Characteristic Dimension (CD) is similar to Average Visibility but only considers the shortest distance. For Dispersion (Dsp), the number of rays is increased to sixteen, and the maximum range is set to a constant value. If a scan ray intersects an occupied region while the previous ray lies in a free region (or vice versa), a change is counted. For these four metrics, the final result is the average over the path. Tortuosity (Trts) is the ratio between the path length and the Euclidean distance between the start and target points. A complete description of these metrics can be found in \cite{BenchmarkingMetricGroundNavigation}.

\subsection{Comparison with Medial-Axis Transform (MA)}
\label{medialAxis}
The medial axis transform ($MA$) \cite{MAPRM}, \cite{choi1997mathematical} is a computational geometry technique used to extract the skeleton of the free workspace. We applied $MA$ to a selection of maps from our dataset to highlight the advantages of our approach over classical methods. The MA transform is used to build a roadmap from a set of random samples $P \in \mathcal{X}_{free}$, such that each sample is retracted onto the medial axis of the free configuration space.

\begin{equation}MA(P) = {x \in P \mid \nexists ; y \in P ; \text{with} ; B_{P}(x) \subseteq B_{P}(y) }\label{eq_medialAxis}\end{equation}

In Equation \eqref{eq_medialAxis}, $B_{P}(\cdot)$ represents the largest closed disc centered at $x$ (or $y$) that is contained in $P$. The results of the MA transform are very similar to the samples obtained from a skeletonization process. Our implementation follows the algorithm described in \cite{MAPRM} with one exception: we always initialize the random samples in the free space to reduce processing time.

\begin{figure}
\centering
\includegraphics[scale=0.48]{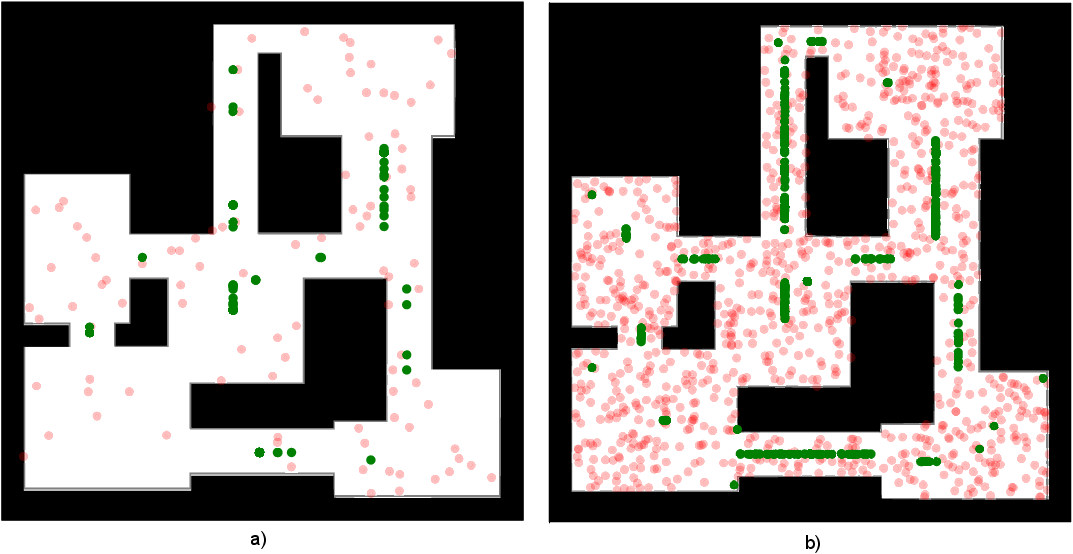}
\caption{Medial-axis transform over an example map with two different rate of samples. In red random samples in free workspace (x). In green the result of MA transform (MA(x))}
\label{rateOfSamples}
\end{figure}

\begin{figure}
\centering
\includegraphics[scale=0.94]{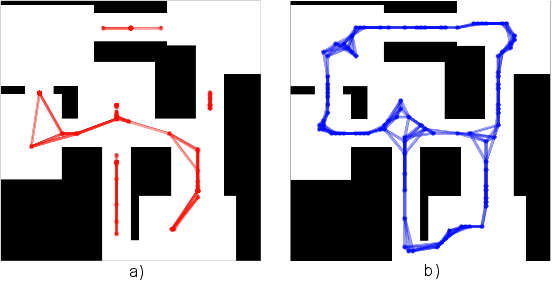}
\caption{roadmap comparative with 100 samples. Left a roadmap applying MA, right roadmap applying SkelUnet.}
\label{fig_roadMapComparative}
\end{figure}

Figure \ref{rateOfSamples} shows how the medial axis, in the context of our dataset, has the disadvantage of retracting samples onto the same specific regions even as we increase the number of samples. Therefore, the resulting roadmap does not cover the entire workspace (see Figure \ref{fig_roadMapComparative}). To connect nodes, we use the $k$-nearest neighbors algorithm with $k=6$. Considering these results, we can identify the need for a more accurate method to obtain samples in safe regions while simultaneously covering the entire free workspace.

\subsection{Analysis} 

From a computer vision perspective, SkelUnet outperforms similar work in skeletonization tasks, demonstrating its effectiveness in processing two-dimensional maps. Based on the experiments, we believe that skeletonized roadmaps have some advantages over Zhang-Suen. Namely, SkelUnet efficiently covers free regions of the workspace without including hard-to-transit areas, showcasing the neural network's efficiency in computing complex path planning tasks.

Analyzing the results from Table \ref{tablePathMetricsUNet} regarding average values, we observe that DTCO has the best performance in paths from Zhang-Suen and SkelUnet, which exhibit very close mean values (a difference of only about $5.2\%$). For A*, the mean is $6.06$, which means that its paths are closer to the obstacles. For the AV index, a small value means the computed path goes through narrow passages, implying greater difficulty for the robot to navigate. The best performance in AV is achieved by Zhang-Suen, followed by SkelUnet with a $2.3\%$ difference.

For Dsp, a low value means that the path follows a clear region. SkelUnet has the best performance, with a difference of $44.0\%$ compared to Zhang-Suen and $368.0\%$ lower than A*, indicating a remarkable difference between the methods. The CD metric is very similar across the three methods because it is a more unspecific metric derived from the AV metric. Finally, regarding tortuosity, a value close to one means short paths. A* has the shortest paths, while Zhang-Suen and SkelUnet only differ by $1.47\%$. The results indicate that SkelUnet and Zhang-Suen generate longer paths than A* but are more feasible for robot navigation. SkelUnet and Zhang-Suen's results are very similar except for Dsp. We can assume that Dsp reflects the diagonal parts of the skeleton removed by SkelUnet. Please note that unlike works such as \cite{Neural_RRT} and \cite{GeneratingCriticalNodesResnet50}, we do not explicitly consider the robot's geometry in the planning scheme because the robot is modeled as a point in the configuration space.

\subsubsection{Scalability and Limitations}

Currently, the method is not suitable for three-dimensional maps, such as voxel maps. The main restriction is that the SkelUnet architecture operates on two-dimensional inputs; therefore, for three-dimensional maps a new architecture is required. Nevertheless, the proposed method was analyzed for two-dimensional maps and tested on a robot with a workspace in $\mathbb{R}^3$. For this purpose, we considered the path as a set of waypoints with constant elevation. This approach demonstrates that the method is suitable for mobile robots. The path computed by our method is used within a global planner, where each vertex represents a position for the robot and the transitions are handled by a local planner.

\section{Conclusions} \label{conclusions}

We presented a method for path planning that constructs a safe roadmap to perform the sampling stage in a single cycle using a neural network called SkelUnet.

According to navigation metrics, SkelUnet-OSS generates more traversable paths than the A* algorithm and demonstrates slight improvements in the dispersion metric. This represents a significant advantage, as it indirectly accounts for uncertainties encountered in real-world robotics applications.

Compared to traditional roadmap methods like the Medial Axis Transform, SkelUnet achieves comprehensive workspace coverage with fewer samples, highlighting its superiority in this particular task. Furthermore, although we employ an edge connection module, if the number of neighbors is low, a graph can be built from SkelUnet without requiring a collision detection stage. The underlying idea of this approach is based on the following assumption: Given a path planning problem, we assume the existence of a path in state space that resolves the query. This route can be expressed in the workspace and therefore serve as a guide for the remaining configuration variables.

In future work, we aim to explore the replacement of collision detection with a learning approach. Another aspect to improve is extending the skeleton version to 3D environments and applying it in more complex tasks.

\section*{Acknowledgments}
This research was funded by Secretaria de Investigaci\'on y Posgrado-IPN grant number 20240705.

\bibliographystyle{IEEEtran}
\bibliography{pathplanning}

\newpage

\section*{Biography Section}

\begin{IEEEbiography}[{\includegraphics[width=1in,height=1.25in,clip,keepaspectratio]{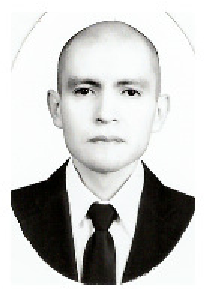}}]{Gabriel O. Flores-Aquino}
received the B.Sc. degree in Mechatronics Engineering in 2017 from the Professional School of Engineering and Advanced Technologies of the National Polytechnic Institute of Mexico (UPIITA-IPN). He received the M.Sc. degree in 2019 and the Ph.D. degree in 2023, both in Advanced Technology from the National Polytechnic Institute in Mexico. He is currently a postdoctoral researcher at the Mathematical Research Center (CIMAT), Guanajuato, Mexico. His research interests include mobile robotics, motion planning, pursuit-evasion problems, and artificial intelligence.

E-mail: gabriel.flores@cimat.mx ORCID: 0000-0001-7951-890X
\end{IEEEbiography}

\begin{IEEEbiography}[{\includegraphics[width=1in,height=1.25in,clip,keepaspectratio]{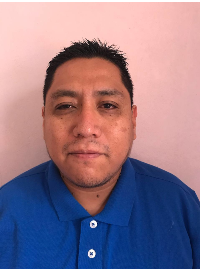}}]{Octavio Gutierrez-Frias}
was born in Mexico City. He received a B.S. degree in Mechatronics from the Professional School of Engineering and Advanced Technologies of the National Polytechnic Institute of Mexico (UPIITA-IPN) in 2003. He received an M.Sc. degree in Computing Engineering from the Computing Research Center of the National Polytechnic Institute in 2006. In 2009, he received a Ph.D. in Computer Sciences at the CIC-IPN. Since 2012, he has been with the Graduate Section at UPIITA-IPN. His research focuses on nonlinear systems control, underactuated systems, robotics, and automation.

E-mail: ogutierrezf@ipn.mx. ORCID: 0000-0002-2855-3243
\end{IEEEbiography}

\begin{IEEEbiography}[{\includegraphics[width=1in,height=1.25in,clip,keepaspectratio]{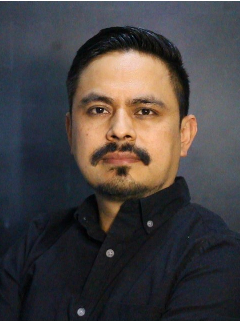}}]{Juan Irving Vasquez}
received his M.Sc. and Ph.D. degrees from the National Institute for Astrophysics, Optics, and Electronics (INAOE), Mexico, in 2009 and 2014, respectively. He earned his B.S. degree in Computer Sciences from the Tehuacan Institute of Technology, Mexico, in 2006. From 2016 to 2021, he served as a researcher at the National Council of Science and Technology (CONACYT) in Mexico. Since 2021, he has been a full-time professor at the National Polytechnic Institute (IPN). His research interests include robotics, motion planning, view planning, and their applications to object reconstruction, inspection, and surveillance.

E-mail: jvasquezg@ipn.mx. ORCID: 0000-0001-8427-9333
\end{IEEEbiography}

\vspace{11pt}

\vfill

\end{document}